\documentclass[letterpaper, 10 pt, conference]{ieeeconf}  
\IEEEoverridecommandlockouts                              
\usepackage[pangram]{blindtext}
\usepackage{amsmath,amsfonts,amssymb}
\let\labelindent\relax
\usepackage{prettyref}
\usepackage{mathrsfs}
\usepackage{graphicx}
\usepackage{wrapfig}
\usepackage{subfig}
\usepackage{MnSymbol}
\usepackage{multirow}
\usepackage{booktabs}
\usepackage{algorithm}
\usepackage[table]{xcolor}
\usepackage[noend]{algpseudocode}
\usepackage{highlight}
\usepackage{enumitem}
\usepackage{comment}
\usepackage{hhline}
\usepackage
[backend=bibtex,
bibstyle=ieee,
citestyle=numeric,
mincitenames=1,
maxcitenames=2,
natbib=true,
doi=false,
isbn=false,
url=false,
eprint=false]{biblatex}
\usepackage[colorlinks,allcolors=gray,hypertexnames=true]{hyperref} 
\newcommand{\citewithauthor}[1]{\citeauthor{#1} \cite{#1}}

\algnewcommand{\LineComment}[1]{\State \(\triangleright\) #1}
\algdef{SE}[DOWHILE]{Do}{doWhile}{\algorithmicdo}[1]{\algorithmicwhile\ #1}
\newcommand*{\colorboxed}{}
\def\colorboxed#1#{%
  \colorboxedAux{#1}%
}
\newcommand*{\colorboxedAux}[3]{%
  \begingroup
    \colorlet{cb@saved}{.}%
    \color#1{#2}%
    \boxed{%
      \color{cb@saved}%
      #3%
    }%
  \endgroup
}

\newrefformat{fig}{Figure~\ref{#1}}
\newrefformat{par}{Section~\ref{#1}}
\newrefformat{appen}{Appendix~\ref{#1}}
\newrefformat{sec}{Section~\ref{#1}}
\newrefformat{sub}{Section~\ref{#1}}
\newrefformat{tab}{Table~\ref{#1}}
\newrefformat{table}{Table~\ref{#1}}
\newrefformat{ass}{Assumption~\ref{#1}}
\newrefformat{alg}{Algorithm~\ref{#1}}
\newrefformat{def}{Definition~\ref{#1}}
\newrefformat{thm}{Theorem~\ref{#1}}
\newrefformat{lem}{Lemma~\ref{#1}}
\newrefformat{step}{Step~\ref{#1}}
\newrefformat{ln}{Line~\ref{#1}}
\newrefformat{eq}{Equation~\eqref{#1}}
\newrefformat{pb}{Problem~\ref{#1}}
\newrefformat{it}{Item~\ref{#1}}
\newrefformat{te}{Term~\ref{#1}}
\def\Eqref Eq:#1:{\eqref{eq:#1}}
\newrefformat{Eq}{Equation~\Eqref#1:}

\newcommand{\TE}[1]{\textbf{#1}}

\newcommand{\FPP}[2]{\frac{\partial{#1}}{\partial{#2}}}

\newcommand{\TWO}[2]{\left(\setlength{\arraycolsep}{1pt}\begin{array}{cc}{#1} & {#2}\end{array}\right)}

\newcommand{\THREE}[3]{\left(\setlength{\arraycolsep}{1pt}\begin{array}{ccc}{#1} & {#2} & {#3}\end{array}\right)}

\newcommand{\argminP}[1]{\text{argmin}}
\newcommand{\argmax}[1]{\underset{#1}{\text{argmax}}}
\newcommand{\argmaxP}[1]{\text{argmax}}
\newcommand{\ST}{\text{s.t.}}

\newcommand{\proofread}[1]{}
\newif\ifArxiv
\usepackage{xcolor}

\usepackage{tikz}
\usetikzlibrary{matrix,calc}
\tikzstyle{block} = [draw,rectangle,thick,minimum height=2em,minimum width=2em]
\tikzstyle{arrow} = [->,thick]
\usepackage{pgfplots}
\pgfplotsset{width=10cm,compat=1.9,every axis plot/.append style={line width=2pt}} 
\makeatletter
\IEEEtriggercmd{\reset@font\normalfont\fontsize{7.0pt}{7.2pt}\selectfont}
\makeatother
\IEEEtriggeratref{1}

\makeatletter
\newcommand\fs@ruled@notop{\def\@fs@cfont{\bfseries}\let\@fs@capt\floatc@ruled
  \def\@fs@pre{}%
  \def\@fs@post{\kern2pt\hrule\relax}%
  \def\@fs@mid{\kern2pt\hrule\kern2pt}%
  \let\@fs@iftopcapt\iftrue}
\renewcommand\fst@algorithm{\fs@ruled@notop}
\makeatother

\title{\Large\bf Learning Reduced-Order Soft Robot Controller}
\author{Chen Liang$^{1,2}$, Xifeng Gao$^{1}$, Kui Wu$^{1}$, Zherong Pan$^{\dagger1}$\\
\thanks{$^\dagger$ indicates corresponding author. $^1$LightSpeed Studios, Tencent (\{xifgao,kwwu,zrpan\}@global.tencent.com). $^2$Department of Computer Science, Zhejiang University.}
\vspace{-30px}}
        
\begin{document}
\maketitle
\thispagestyle{empty}
\pagestyle{empty}

\begin{abstract}
Deformable robots are notoriously difficult to model or control due to its high-dimensional configuration spaces. Direct trajectory optimization suffers from the curse-of-dimensionality and incurs a high computational cost, while learning-based controller optimization methods are sensitive to hyper-parameter tuning. To overcome these limitations, we hypothesize that high fidelity soft robots can be both simulated and controlled by restricting to low-dimensional spaces. Under such assumption, we propose a two-stage algorithm to identify such simulation- and control-spaces. Our method first identifies the so-called simulation-space that captures the salient deformation modes, to which the robot's governing equation is restricted. We then identify the control-space, to which control signals are restricted. We propose a multi-fidelity Riemannian Bayesian bilevel optimization to identify task-specific control spaces. We show that the dimension of control-space can be less than $10$ for a high-DOF soft robot to accomplish walking and swimming tasks, allowing low-dimensional MPC controllers to be applied to soft robots with tractable computational complexity.
\end{abstract}
\section{Introduction}
\begin{figure*}[th]
\centering
\begin{tabular}{c}
\includegraphics[trim=0 12cm 0 12cm,clip,width=.95\linewidth]{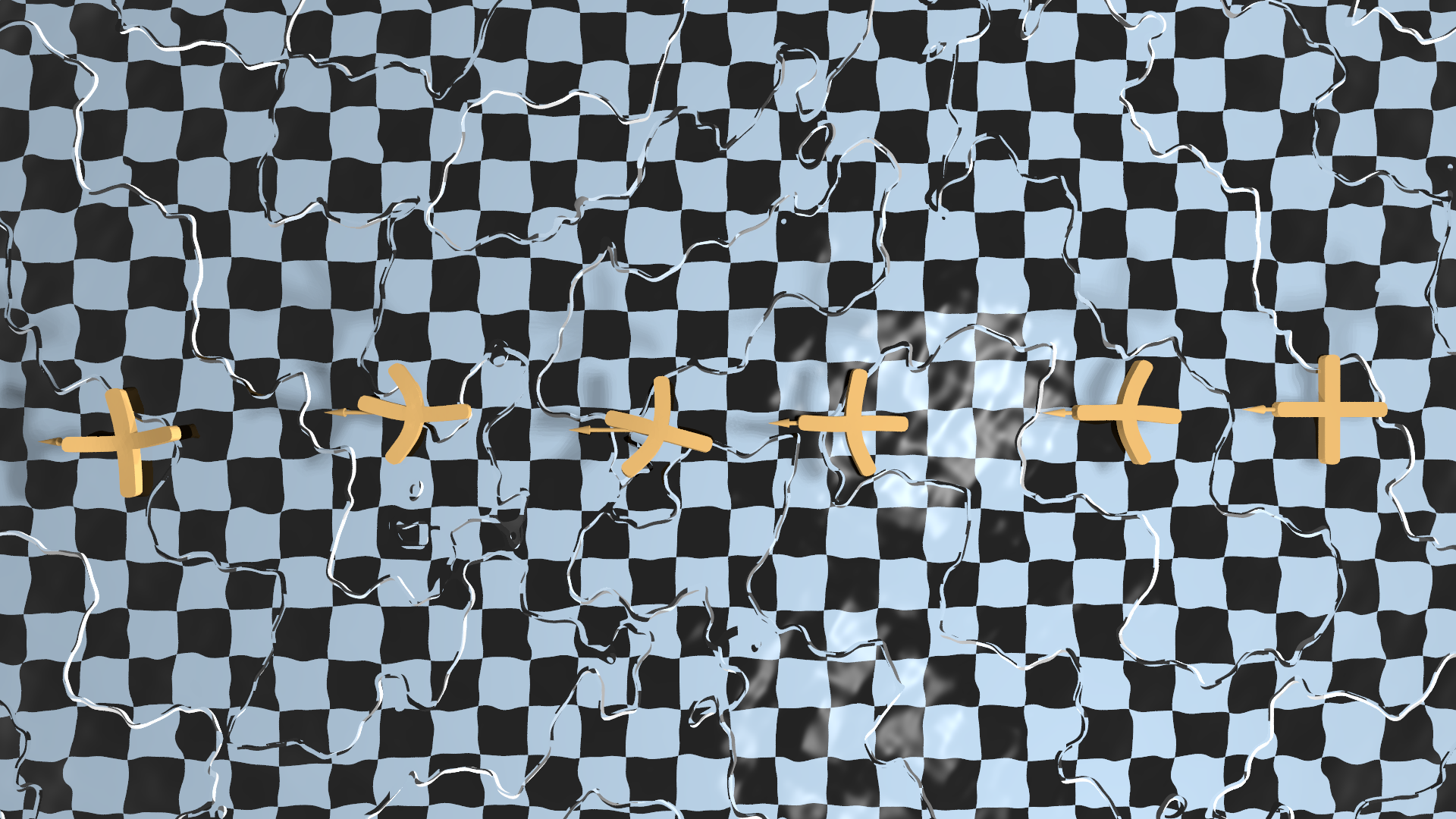}
\put(-25,20){(a)}\\
\includegraphics[trim=0 12cm 0 12cm,clip,width=.95\linewidth]{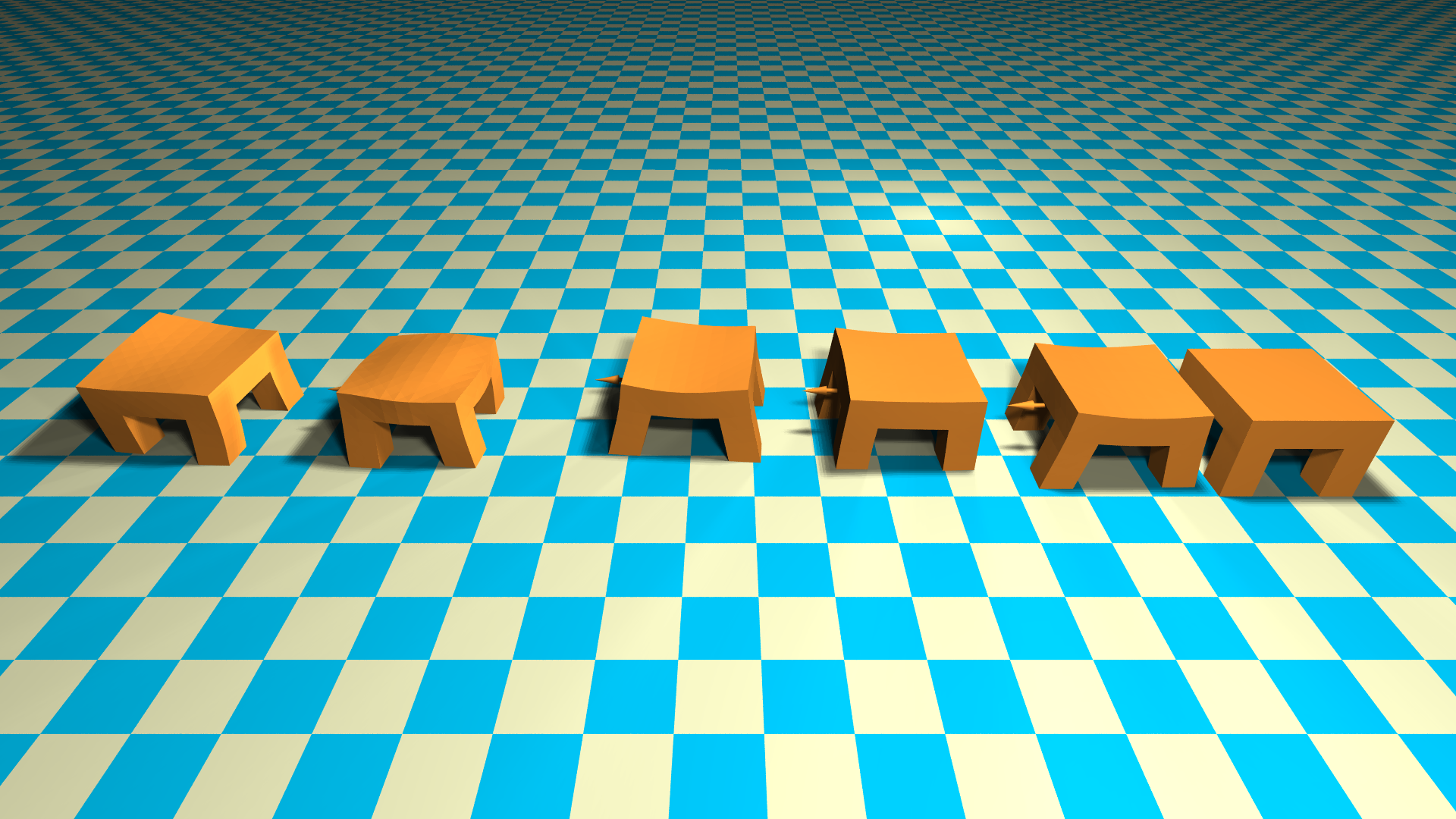}
\put(-25,20){(b)}
\end{tabular}
\caption{\label{fig:tasks} Two locomotive tasks considered in our evaluation: A soft cross swimming (a); A soft quadruped walking (b), where we show robot poses at different time instances.}
\vspace{-10px}
\end{figure*}
As compared with rigid structures, soft materials pertain a higher flexibility and a lower manufacturing cost (we refer readers to~\cite{polygerinos2017soft} for a thorough overview). Indeed, the rigidity of materials can significantly limit the mode of deformation, so articulated robots oftentimes require precision servo motors for actuation. Instead, soft robots utilize the material compliance to conduct forces and induce deformations, which can be controlled using low-cost pneumatic or cable-based actuators. Over the years, we have witnessed soft robots exhibit superior flexibility in certain manipulation tasks including universal object grasping~\cite{amend2012positive} and gait-adaptive navigation~\cite{shepherd2011multigait}. However, the number of robotic tasks accomplished by soft robots is still incomparable to those accomplished by conventional articulated robots, which is due to a lack of effective soft robot control techniques. Sadly, almost all existing soft robot hardware platforms rely on manually designed gaits to accomplished specified tasks, while articulated robots can utilize a row of general-purpose, automatic planning and control algorithms such as rapid exploring random trees (RRT) and model-predictive controllers (MPC) controllers.

To design an effective soft robot controller, an algorithm has to conquer the curse-of-dimensionality. Indeed, classical continuum theory~\cite{fung2017classical} of elasticity assumes that every infinitesimal soft tissue can deform independently and its configuration space is infinite-dimensional. Modern computational models, e.g. the finite element method (FEM), discretize the configuration space into a finite-dimensional functional space. However, the resulting dimension of the discrete space can be at the level of hundreds or even thousands~\cite{pan2018realtime,bern2019trajectory,hu2019chainqueen}. Although conventional planning and control algorithms can be adopted after the discretization, their computational cost is prohibitively high. For example, the cost of MPC controller grows cubically~\cite{tassa2012synthesis} and that of the optimal RRT algorithm grows exponentially~\cite{janson2018deterministic} as the increase of the dimensions in the configuration space.

On a parallel front, efforts have been made to design automatic control algorithms for soft robots, which can be classified into trajectory optimization techniques, shooting methods, and learning-based algorithms. Trajectory optimization techniques~\cite{pan2018active,bern2019trajectory} perform long-horizon planning by directly optimizing the pose of the soft robot at sampled time instances. As a result, their dimension of search space is at least tens of thousands, leading to expensive computations. Similarly, shooting methods~\cite{li2019propagation,spielberg2019learning} formulate the receding-horizon control problem as a local nonlinear optimization, which can be solved via gradient-based methods. These gradient information can be provided by learned~\cite{li2019propagation} or analytically derived~\cite{spielberg2019learning} differentiable dynamic models. However, the computational cost of such gradient evaluation is still polynomial in the dimension of the configuration space. Finally, learning-based algorithms~\cite{thuruthel2018model,pan2018realtime,bhatia2021evolution} represent the controller (or a part of the controller) using a neural network, which is then trained via reinforcement learning (RL). These methods have achieve an unprecedented level of success on navigation tasks, but they require excessive tuning of RL hyper-parameters and reward signals. 

\TE{Main Results:} Inspired by the success of Reduced-Order Modeling (ROM)~\cite{pan2018active}, we propose a novel controller synthesis method based on the low-dimensional assumption. Specifically, we assume that, in order to faithfully model soft robots, we only need to restrict its configuration space to a low-dimensional subspace that captures the salient deformation modes, which is denoted as the simulation-space. We further assume that, soft robots can accomplish locomotion tasks via under-actuation, i.e., restricting the space of control signals to a subspace of an even lower dimension, which is denoted as the control-space. We identify these crucial spaces in two stages. First, we use a conventional model analysis technique~\cite{an2008optimizing} to identify the linear simulation-space and construct the restricted dynamic system. We then identify the bases of the control-space as a subset of the simulation-space bases. To this end, we propose a multi-fidelity Riemannian Bayesian Bilevel Optimization (RBBO) technique. Our low-level optimizer is an MPC controller, which maximizes the reward function of the task, while our high-level Bayesian optimizer selects the control-space bases to maximize the performance of low-level controller. Put together, our RBBO algorithm can automatically discover the most effective deformation modes to accomplish a locomotion task.

We have evaluated our method on the walking and swimming tasks of several soft robots. Our results show that RBBO can significantly improve the controller performance while restricting the dimension for control space to be less than $10$. Thanks to such a low dimension, both soft robot simulation and restricted MPPI or MPC control signals can be computed at a reasonable cost.

\section{Related Work}
We consider soft robots as those made out of flexible compliant materials. We review related works in soft robot design, modeling, and control, with a focus on reduced-order techniques.

\subsection{Design} 
Unlike articulated robots where a general-purpose robot design can be used to accomplished many tasks, existing soft-robots are still heavily engineered towards one type of tasks. Three tasks have been actively studied: positioning \& tracking~\cite{doi:10.1177/0278364915587925}, locomotion~\cite{shepherd2011multigait}, and grasping~\cite{amend2012positive}. Various soft robot arms has been designed to accomplish end-effector positioning and tracking tasks, including cable-driven~\cite{wang2013visual} or pneumatic~\cite{doi:10.1177/0278364915587925} multi-segment soft structures. However, to accomplish more challenging manipulation or locomotion tasks, such as walking and grasping, soft robots must be designed to interact with the environment. \citewithauthor{shepherd2011multigait} proposed a bio-inspired pneumatic crawling robot with multi-chamber leg-like structures. \citewithauthor{bern2019trajectory} proposed a cable-driven soft walking robot and used trajectory optimization to automatically search for walking gaits. Several more recent works~\cite{cheney2014unshackling,bhatia2021evolution} show that many soft-robot designs can achieve equally effective walking performance and evolutionary algorithms can be used to automatically search for such designs. Finally, many works have advocated grasping as a potential application of soft robot arms. However, little control can be applied to the grasping procedure, since most soft grippers~\cite{amend2012positive,giannaccini2014variable} only have one degree of freedom. These methods are orthogonal to our work, which is focused on soft robot modeling \& control. We will show that our method can be applied to robots of arbitrary shape and modality.

\subsection{Modeling}
Fast articulated robot simulation is a well-established tool for robot design validation and model-based planning \& control. However, universally high-performance simulation is still unavailable to soft robots due to the prohibitive computational overhead caused by high-dimensional configuration spaces. A row of task- and design-specific kinematic and dynamic soft robot models have been adopted. For multi-segment soft robot arms~\cite{renda2016discrete} and steerable needles~\cite{chentanez2009interactive}, the Cosserat theory can be exploited to model robot as a thick-rod with twisting and bending degrees of freedom. Some soft robots consist of thin shell-like structures and can be simulated using membrane dynamic models~\cite{huang2020dynamic}. The vast majority of simulation tools, including commercial softwares like Ansys and COMSOL, are based on FEM (see~\cite{xavier2021finite} for a thorough survey), incorporating various discretization methods and material models. For example, \citet{fang2018geometry} modeled cable-driven soft arm via ARAP deformable model to enable fast simulation via global-local solvers. \citet{hu2019chainqueen} simulated crawling robots using material point methods with hybrid particle-grid representation. \citet{cheney2014unshackling} uses a mass-spring-damper system to simulate heterogeneous robots. All these FEM variants incur a computational cost that is superlinear in the dimension of configuration spaces.

ROM can significantly boost the computational performance of soft-robot simulation by restricting the configuration space to a low-dimensional linear or nonlinear manifold. These methods have been widely adopted to the modeling of fluid~\cite{carlberg2013gnat} and deformable objects~\cite{hauser2003interactive}. Recently, these methods have been gradually used to model articulated and soft robots. \citet{chen2020optimal} searched for ROM that is best suited for legged robot locomotive tasks. \citet{pan2018active} used ROM as physics constraints in trajectory optimization for soft robots. \citet{sadati2019reduced} proposed to use ROM for simulating soft continuum manipulators. In comparison, our method follows these works and used ROM for soft robot dynamics. However, we use two separate subspaces for simulation and control, where the simulation-space is analytically determined, while the control space is automatically optimized.

\subsection{Control}
Gaits of most manually designed soft robots~\cite{amend2012positive,shepherd2011multigait} are also hand-engineered and the underlying controller only need to track the designed gaits. The automatic controller design for soft robots emerge very recently. \citet{cheney2014unshackling} assume robots consist of several classes of volume-controllable blocks, whose locations are optimized to best perform the walking task, but their controller is open-loop. A prominent closed-loop controller is the receding-horizon shooting method such as MPPI~\cite{theodorou2010generalized} and iLQR~\cite{tassa2012synthesis}. However, these algorithms have not been largely used to control soft robots, because their computational cost scales superlinearly with the dimension of configuration space. Several methods have been proposed to overcome this challenge. \citet{hu2019chainqueen} proposed a differentiable soft robot simulator allowing gradient-based solution of associated optimization problem of the shooting method. \cite{thuruthel2018model} used reinforcement learning to optimize a parametric controller. However, the performance of reinforcement learning is sensitive to both hyper-parameters and controller representations. In parallel, \citet{pan2018active} showed that using ROM can effectively reduce the cost of trajectory optimization for soft robots. However, the degree of freedom (DOF) induced by ROM is still larger than $50$ for a typical soft robot, making is too costly to apply shooting method. In this work, we further apply underactuation and restrict the control signals to a control-space of up to 5-dimensional, allowing MPC to be efficiently applied to find soft robot walking and swimming gaits.
\section{Problem Statement}
In this section, we formulate the problem of soft robot control. We will frequently deal with restricted configuration space of a soft robot. A soft robot takes up a volume $\Omega_0$ when no external forces are exerted, which is known as the rest shape, and we use $x\in\Omega_0$ to denote a continuous point in the rest domain. Under external forces, it deforms to take a volume $\Omega$ under the deformation function $\phi(x,t): \Omega_0\mapsto\Omega$. At the same time, a potential energy functional $\mathcal{P}[\phi]$ is induced by $\phi$ as: 
\begin{align*}
\mathcal{P}[\phi]\triangleq\int_{\Omega_0}\psi(\nabla\phi)dx,
\end{align*}
where $\psi$ is the potential energy density. The dynamics of a soft robot is governed by the Euler-Lagrangian dynamics associated with the following Lagrangian function:
\begin{align*}
\mathcal{L}[\phi]=\frac{\rho}{2}\int_{\Omega_0}\|\dot{\rho}\|^2dx-\mathcal{P}[\phi],
\end{align*}
where $\rho$ is the density of robot. We denote the infinite dimensional dynamic system as an functional:
\begin{align*}
\ddot{\phi}=f[\phi,\dot{\phi}].
\end{align*}
We simulate the soft robot using finite element method, where $\phi$ is discretized using a volumetric mesh with $N$ nodes, and the vector of node positions is denoted as $q\in\mathbb{R}^{3N}$. Given $q$, the continuous function $\phi$ is approximated as: $\phi(x,t)=B(x,t)q$, where $B(x,t)\in\mathbb{R}^{3\times 3N}$ is the vector of finite-element shape functions. Plugging this approximation into the weak-form dynamic system $f$, a finite-dimensional discrete system can be represented as the following function (see~\cite{fung2017classical,xavier2021finite} for its detailed derivation):
\begin{align*}
\ddot{q}=f(q,\dot{q}).
\end{align*}
Being a computable model, the volumetric mesh typically involves tens of thousands of nodes leading to a high cost in time-integrating $q$ via $f$. 

\subsection{Reduced-Order Modeling}
ROM is an useful tool that allows users to discover salient global deformation modes of a soft robot, while ignoring small-scale local deformations. ROM also enables efficient robot simulation by focusing the computation only on the salient modes. A key difference between FEM and ROM lies in the features of bases $B$. The FEM utilizes locally supported $B$, where each element of $B$ is non-zero only within a small neighborhood around one node. Instead, ROM assumes globally supported $B$ so that an element can represent a global deformation, allowing ROM to use a small number of bases to represent salient deformation modes. 

Although we could unify the theory of FEM and ROM by replacing the bases, ROM is typically built on top of FEM. This is because ROM relies on a reasonable set of bases to capture the salient deformation modes, which are typically derived by analyzing the deformation patterns of the FEM system~\cite{hauser2003interactive,barbivc2005real,an2008optimizing}. Therefore, we formulate ROM as a linear subspace of $B$, denoted as the $M$-dimensional simulation-space with linear bases matrix $B_r\in\mathbb{R}^{3N\times M}$ and $M\ll N$. The configuration space of ROM is thus only $M$-dimensional and the reduced configuration $q_r$ is related to $q$ via $q=B_rq_r$. Various techniques have been proposed~\cite{hauser2003interactive,barbivc2005real,an2008optimizing} to further restrict the dynamic system $f$ to the simulation-space and we denote the restricted function as:
\begin{align}
\label{eq:ROM}
\ddot{q}_r=f_r(q_r,\dot{q}_r).
\end{align}
The ROM can be time-integrated at a much lower cost. However, \prettyref{eq:ROM} is still too costly to be used as a predictive model for MPC. This is because the reduced dimension $M$ can still be larger than $50$ as used by prior works~\cite{barbivc2005real,an2008optimizing,pan2018active}, which is much larger than conventional articulated robots. Indeed, during each control loop with horizon $H$, a MPC controller needs to perform $HM$ evaluations of $f_r$ in order to compute the state derivatives.

\subsection{Contact and Actuation}
In order for the soft robot to be actuated and interact with environment, we follow prior work~\cite{pan2018active} and introduce both internal force $f_r$ and external force $f_e$, leading to the following forced dynamic system:
\begin{align}
\label{eq:ROMControl}
\ddot{q}_r=f_r(q_r,\dot{q}_r)+M_r^{-1}(f_e+f_r),
\end{align}
where $M_r$ is the reduced mass matrix. The external force $f_e$ is computed via the nonlinear complementary problem (NCP). Specifically, for the $i$th FEM mesh node located at $B(x_i,t)B_rq_r$ with rest state position being $x_i$, we detect any collisions between the node and environment. If a collision is detected with contact normal $n_i$, then a contact force $f_i$ is applied on the $i$th node. To determine $f_i$, we introduce the following complementary constraint in reduced coordinates:
\begin{equation}
\begin{aligned}
\label{eq:LCPConstraint}
&f_e\triangleq\sum_{x_i\text{ in contact}}B_r^TB(x_i,t)^Tf_i\\
&\dot{q}_{i,\perp}\triangleq n_i^TB(x_i,t)B_r\dot{q}_r\\
&\dot{q}_{i,\parallel}\triangleq (I-n_in_i^T)B(x_i,t)B_r\dot{q}_r\\
&0\leq \dot{q}_{i,\perp}\perp n_i^Tf_i\geq0\\
&0=\dot{q}_{i,\parallel}+\lambda\frac{(I-n_in_i^T)f_i}{\|(I-n_in_i^T)f_i\|}\\
&0\leq \lambda\perp \mu n_i^Tf_i-\|(I-n_in_i^T)f_i\|\geq0,
\end{aligned}
\end{equation}
where $\mu$ is the frictional coefficient and $\lambda$ is the tangential force coefficient. The system of equation given by~\prettyref{eq:ROMControl} and~\prettyref{eq:LCPConstraint} can be solved using Newton's method to yield $f_e$ and time integrate $q_r$. However, owning to the dense FEM mesh, the number of contacts can be large and exactly solving the NLP problem is intractable. Instead, we use the staggered projection~\cite{KSJP:2008} as an approximate solver, which alternatively updates the normal and tangential component of $f_e$ via two convex programs.

\subsection{Locomotion Control}
The internal forces $f_i$ is actuator generated and the main goal of our method is to design an algorithm that can automatically and efficiently compute $f_i$ for the robot to accomplish various locomotion tasks. Each locomotion task is described by a reward function $R(q,\dot{q})$ so that MPC can be applied to solve the following optimization:
\begin{equation}
\begin{aligned}
\label{eq:OPT}
\argmax{f_r^n}\;&\sum_{n=1}^HR(q^{n+1},\dot{q}^{n+1})\\
\ST\quad&\frac{\dot{q}_r^{n+1}-\dot{q}_r^n}{\Delta t}=f_r(q_r^{n+1},\dot{q}_r^{n+1})+M_r^{-1}(f_e+f_r^n)\\
&\dot{q}_r^{n+1}\triangleq\frac{q_r^{n+1}-q^n_r}{\Delta t},
\end{aligned}
\end{equation}
leading to the standard receding-horizon feed-back control algorithm. Here we use superscripts to denote timestep index and $\Delta t$ is the timestep size. However, the na\"ive application of existing MPC algorithms such as~\cite{tassa2012synthesis} would involve at least $HM$ calls to the simulator for computing the state derivatives $\FPP{q_r^{n+1}}{\TWO{q_r^n}{\dot{q}_r^n}}$ via finite difference, leading to prohibitive computational overhead.
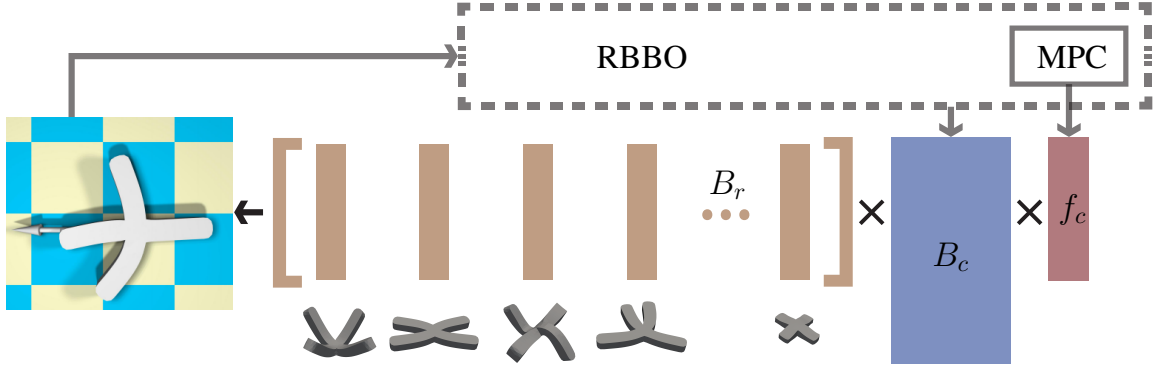
\begin{figure*}[th]
\centering
\scalebox{.9}{
\includegraphics[width=\linewidth]{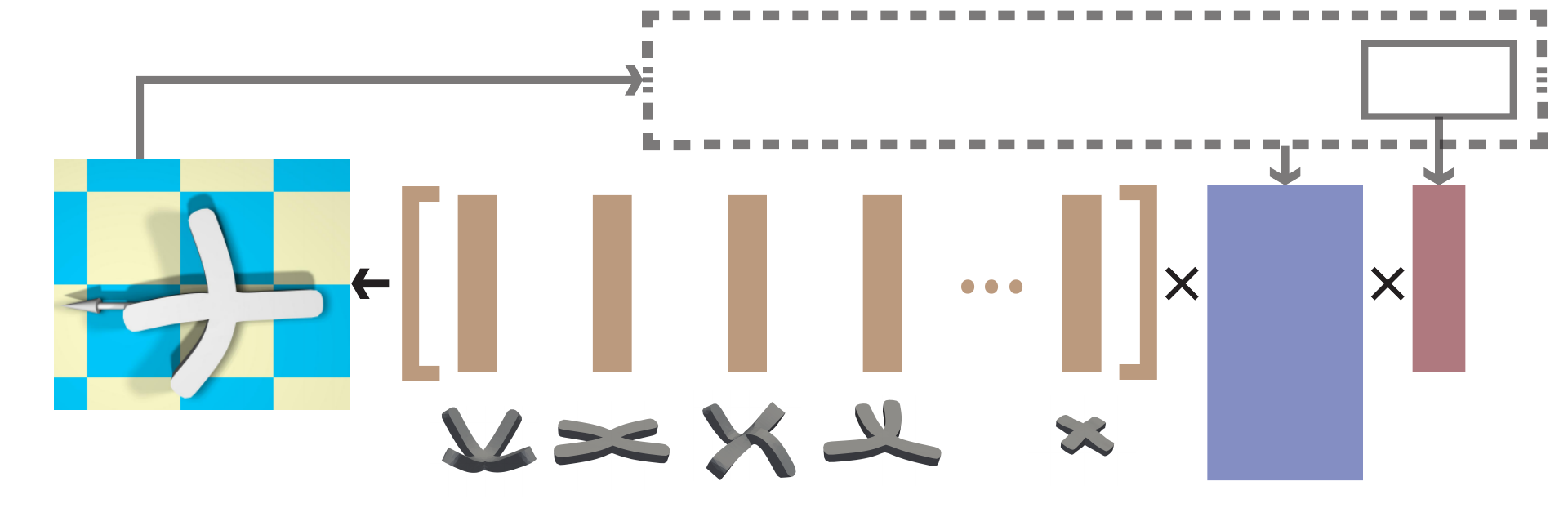}
\put(-193,80){\Large $B_r$}
\put(-99,50){\Large $B_c$}
\put(-46,70){\Large $f_c$}
\put(-240,133){\Large RBBO}
\put(-55,133){\Large MPC}}
\caption{\small{We illustrate the pipeline of our controller optimization algorithm for deformable objects. The equation of motion is restricted to a low-dimensional space spanned by bases $B_r$ (brown). The control signal $f_c$ (red) is further restricted to a lower-dimensional control space spanned by $B_c$ (blue). $B_r$ is constructed analytically to capture the salient deformations as illustrated below each column, while $B_c$ is optimized to maximize the performance of the MPC~\prettyref{eq:OPT}. The optimization is accomplished via the proposed RBBO~\prettyref{alg:RBBO}.}}
\label{fig:pipeline}
\vspace{-10px}
\end{figure*}
\section{Reduced-Order Controller Design}
Our controller design and optimization scheme is illustrated in~\prettyref{fig:pipeline}. Our goal is to design an efficient controller that utilizes the low-dimensional nature of ROM. We observe that many soft robots are highly redundant and under-actuated. Therefore, we propose to further under-actuate the ROM system by introducing an even lower-dimensional control-space. We assume the control-space is a linear subspace of $B_r$ that is specified by the bases matrix $B_c\in\mathbb{R}^{M\times C}$ with $C\ll M\ll 3N$. Therefore, the control force $f_c\in\mathbb{R}^C$ is related to $f_r$ by the linear relationship: $f_r=B_cf_c$. The design philosophy of control-space follows the same idea as that of the simulation-space. Given an arbitrary robot, ROM provides an automatic tool to discover the salient deformation modes encoded in the globally supported bases $B_r$. When we are further given a locomotion task in the form of a reward function $R(q,\dot{q})$, our method provides an automatic tool to discover the subspace of control signals, encoded in the bases $B_c$, that can most effectively accomplish the task. 

\subsection{MPC in Control-Space}
The bottleneck of applying iLQR~\cite{tassa2012synthesis} to ROM lies in the $HM$ calls to the simulator. Unfortunately, although our control-space reduces the dimension of control signals, the number of calls to the simulator is still at least $H(M+C)$ times for evaluating the state derivatives $\FPP{q_r^{n+1}}{\TWO{q_r^n}{\dot{q}_r^n}}$ and control derivatives $\FPP{q_r^{n+1}}{f_c^n}$ for every $n=1,\cdots,H$. To utilize the control-space and remove the dependency on $M$, we directly compute the sensitivity of the trajectory with respect to the control signal. Let us define the following shorthand notation:
\begin{align*}
Q_r\triangleq&\THREE{q_r^2}{\cdots}{q_r^{H+1}}^T\\
F_c\triangleq&\THREE{f_c^1}{\cdots}{f_c^H}^T\\
\mathcal{R}(Q_r)\triangleq&\sum_{n=1}^HR(q^{n+1},\dot{q}^{n+1}).
\end{align*}
The Gauss-Newton method only evaluates the Jacobian matrix $\nabla_{F_c}Q_r$ and uses the following rule to update the control signal via the Newton-type iteration:
\begin{align*}
F_c\gets F_c-\alpha
(\nabla_{F_c}Q_r^T\nabla_{Q_r}^2\mathcal{R}\nabla_{F_c}Q_r)^{-1}
(\nabla_{Q_r}\mathcal{R}\nabla_{F_c}Q_r)^T,
\end{align*}
where $\alpha$ is a step size parameter and we use the Gauss-Newton approximation for the Hessian matrix. In addition, the control signals cannot change abruptly, so we can further regularize the control signal using a $K-1$th order spline curve with $K$ control points denoted as: $\tilde{F}_c\triangleq\THREE{\tilde{f}_c^1}{\cdots}{\tilde{f}_c^K}^T$. The control signal is then linearly related to $\tilde{F}_c$ as $F_c=S\tilde{F}_c$ via an spline interpolation matrix $S\in\mathbb{R}^{HC\times KC}$. The Gauss-Newton method in this case only requires the Jacobian $\nabla_{\tilde{F}_c}Q_r$ leading to only $HKC$ simulator calls. In practice, this implies sampling $KC$ threads in parallel. Another widely used MPC controller is MPPI~\cite{theodorou2010generalized} and we can adopt MPPI in our control-space by sampling $KC$ trajectories and compute $\tilde{F}_c$ from the reward-weight averaging of their control signals.

\subsection{Bilevel Control-Space Optimization}
The choice of control-space $B_c$ is crucial to the performance of the MPC algorithm. Automatically optimizing $B_c$ is much more difficult than choosing the simulation-space $B_r$. This is because prior works~\cite{hauser2003interactive,barbivc2005real,an2008optimizing} have shown that $B_r$ can be chosen by analyzing the potential energy. However, the control-space can affect the entire optimization process described in~\prettyref{eq:OPT}, which is in turn related to the behavior of the contact-rich dynamic system over an entire control horizon. As a result, we cannot use the gradient-based algorithm to optimize $B_c$ as the contact mechanism is non-differentiable and some MPC algorithm is stochastic. Fortunately, since we assume the control-space is a simulation-subspace, the size of $B_c$ is rather small. Indeed, we can assume that $B_c$ is always an orthogonal matrix so that all the valid $B_c$ lies in the Stiefel Manifold $\text{St}(M,C)$ and since the order of bases can be aribitrary, we have $B_cB_c^T$ lies in the lower-dimensional Grassmannian manifold $\text{Gr}(M,C)$. We formulate this challenging problem as the following bilevel optimization:
\begin{align*}
&\argmax{B_c}\;\mathcal{R}(Q_r)\\
&\ST\quad B_cB_c^T\in \text{Gr}(M,C)\\
&F_c:\begin{cases}
&\argmax{\tilde{F}_c}\;\mathcal{R}(Q_r)-\Lambda(Q_r,\tilde{F}_c)\\
&\ST\;\frac{\dot{q}_r^{n+1}-\dot{q}_r^n}{\Delta t}=f_r(q_r^{n+1},\dot{q}_r^{n+1})+M_r^{-1}(f_e+B_cf_c^n).
\end{cases}
\end{align*}
Note that the goal of our high- and low-level problem is slightly different. The high-level solver aims at maximizing the performance of MPC, which is described by the reward function $\mathcal{R}(Q_r)$. Our low-level MPC controller not only solves the task but also ensures that the dynamic system is stable and the control scheme is energy-efficient, so we introduce the additional regularization term $\Lambda(Q_r,\tilde{F}_c)$.

\subsection{RBBO Algorithm}
Solving the above bilevel optimization is computationally challenging. Even a single call to the low-level problem would involve running MPC over an entire trajectory of $T$ timesteps. We propose to adopt Bayesian optimization~\cite{jaquier2020bayesian} that utilizes the smoothness of objective function with respect to $B_c$. Given a dataset $\mathcal{D}=\{<B_c^i,\mathcal{R}^i>\}$ of past calls to the low-level problem, the Bayesian optimization assumes that the function $\mathcal{R}(B_c)$ follows a Gaussian process:
\begin{align*}
\mathcal{R}(B_c)\sim\text{GP}(B_c;\mathcal{D}).
\end{align*}
Using the Gaussian process, the optimizer can predicts the MPC performance at any $B_c$ as a normal distribution with mean $\mu(B_c)$ and covariance $\sigma(B_c)$, which is also conditioned on the choice of a kernel function $k(B_c^i,B_c^j,\theta)$. The kernel function determines the similarity between two choices of $B_c$, where $\theta$ is some hyper-parameters. Most kernel can also be written as a function $k(d(B_c^i,B_c^j))$ with $d$ being some distance metric. Since we merely require $B_cB_c^T\in \text{Gr}(M,C)$, the Euclidean distance between $B_c^i$ and $B_c^j$ is not a valid similarity measure. The idea of Riemannian Bayesian optimization~\cite{jaquier2020bayesian} lies in the use of Geodesic distance (an intrinsic metric) for $d$, denoted as $d_\mathcal{M}$. Although the geodesic distance on the Grassmannian manifold is feasible to compute~\cite{wen2013feasible}, recent research~\cite{lin2019extrinsic} argues that intrinsic metrics can violate the positive-definiteness of the kernel and advocates the use of extrinsic metrics. Following this observation, we embed $\text{Gr}(M,C)$ into the $M\times M$ ambient space and use the Euclidean distance measure:
\begin{align*}
d(B_c^i,B_c^j)\triangleq\|B_c^i(B_c^i)^T-B_c^j(B_c^j)^T\|,
\end{align*}
and we use the radial kernel for $k$. Based on the prediction of $\mathcal{R}(B_c)$, Bayesian optimizer chooses the next point by maximizing the acquisition function, where we use the GP-UCB function:
\begin{align*}
\gamma(B_c)\triangleq\mu(B_c)+\beta^{1/2}\sigma(B_c),
\end{align*}
where the first term exploits existing knowledge and the second term encourages exploration, weighted by an appropriate parameter $\beta$. We maximize $\gamma(B_c)$ on the Grassmannian manifold via the Riemannian-quasi-newton algorithm~\cite{yuan2016riemannian}.

\begin{algorithm}
\caption{\label{alg:RBBO} RBBO}
\begin{algorithmic}[1]
\Require{ROM dynamic model: $B_r,f_r,M_r$}
\Require{MPC module: MPPI or iLQR}
\Require{Task reward: $R$}
\Ensure{$B_c$}
\State Sample initial dataset $\mathcal{D}'$
\State Fit $\text{GP}$ from $\mathcal{D}'$
\For{$i=1,2,\cdots,N$}
\State $B_c^i\gets\argmax{B_c}\gamma(B_c)\;\ST\;B_cB_c^T\in \text{Gr}(M,C)$
\State Use BOCA to find $T^i$
\State Use MPC to generate a $T^i$-timestep trajectory
\State $\mathcal{D}'\gets\mathcal{D}'\bigcup\{<B_c^i,T^i,\mathcal{R}^i>\}$
\State Fit $\text{GP}$ from $\mathcal{D}'$
\EndFor
\State Return $B_c^i$ with $i=\argmax{i}\;\mathcal{R}^i$
\end{algorithmic}
\end{algorithm}
To further accelerate the optimization, we observe that, in order to confirm a certain choice of $B_c$ is ``bad'', we do not need to run the entire $T$-timestep trajectory. Oftentimes, a bad $B_c$ will lead to sub-optimal performance of MPC at first few timesteps, in which case we can terminate the trajectory to save computation. This observation has been utilized in Bayesian optimization~\cite{kandasamy2017multi} by introducing the multi-fidelity mechanism. Specifically, we treat $T$ as a continuous fidelity level parameter, i.e. treating $T$ as an additional parameter when calling the low-level problem. Using a high-fidelity estimation is more expensive but leads to more accurate result, while a low-fidelity estimation is less expensive and accurate, but provide information about results under higher fidelity due to smoothness. To utilize such information from low-fidelity estimation, RBBO incorporate the BOCA algorithm~\cite{kandasamy2017multi}. Given an augmented dataset $\mathcal{D}'=\{<T^i,B_c^i,\mathcal{R}^i>\}$ with varying fidelity level, RBBO assumes the function $\mathcal{R}(B_c,T)$ follows a joint Gaussian process:
\begin{align*}
\mathcal{R}(B_c,T)\sim\text{GP}(B_c,T;\mathcal{D}'),
\end{align*}
where we use the separable kernel function: $k(B_c^i,T^i,B_c^j,T^j)=k(d(B_c^i,B_c^j))k(|T^i-T^j|)$. After finding the next evaluation point $B_c$, RBBO uses BOCA to select the next fidelity level $T^{i+1}$. The complete RBBO is summarized in~\prettyref{alg:RBBO}.
\section{Experiments}
\setlength{\tabcolsep}{3pt}
\begin{table}[ht]
\centering
\begin{tabular}{cccccccc}
\toprule
Robot & Task & $N$ & $M$ & $C$ & $\mathcal{R}_\text{init}$ & $\mathcal{R}^\star$ & $\mathcal{R}^\star/\mathcal{R}_\text{init}-1$\\
\midrule
Cross & Swim & 623 & 20 & 3  & 2.24 & 5.43 & 142.21\%\\ 
Cross & Walk & 623 & 20 & 3  & 0.66 & 4.28 & 544.32\%\\ 
Beam & Walk & 426 & 20 & 3  & 1.34 & 4.52 & 236.91\%\\ 
Quadruped & Walk & 1315 & 20 & 3  & 1.86 & 5.26 & 183.35\%\\ 
Tripod & Walk & 1417 & 20 & 3  & 2.16 & 3.03 & 40.66\%\\ 
\bottomrule
\end{tabular}
\caption{\small{From left to right: name of robot, locomotion task, full space dimension, simulation subspace dimension, control subspace dimension, the performance of identity baseline $\mathcal{R}_\text{init}$, the performance of RBBO-optimized controller, and improvement.}}
\label{tab:results}
\vspace{-10px}
\end{table}
We perform a series of comparative and ablation studies to demonstrate the effectiveness of our controller design and optimization scheme (refer to our video for visualization). We implement our deformable object simulator in C++ and we generate multiple trajectories in parallel under perturbed control signals, which is the bottleneck of our MPPI or iLQR algorithm. The RBBO algorithm runs in Python and we use the Pymanopt library~\cite{JMLR:v17:16-177} to optimize the acquisition function over the Stiefel manifold. All experiments are performed on a single server with a 48-core AMD EPYC 7K62 CPU. As illustrated in~\prettyref{fig:tasks}, we evaluate our method on two locomotive tasks, walking and swimming, using a set of soft robots with drastically different modalities and degrees of freedom as summarized in~\prettyref{tab:results}. For the swimming task, we model the fluid drag forces using the heuristic force model proposed in~\cite{pan2018active}. In both tasks, our reward function is the direction moving distance, i.e., $R(q^{n+1},\dot{q}^{n+1})=d^T\dot{q}^{n+1}\Delta t$, where $d$ is the desired moving direction. Further, our low-level regularization term $\Lambda(Q_r,\tilde{F}_c)$ penalizes robot orientation changes and any motions orthogonal to $d$. $\Lambda(Q_r,\tilde{F}_c)$ also involves a small control regularization term.

\begin{figure}[ht]
\centering
\scalebox{.7}{
\includegraphics[width=\linewidth]{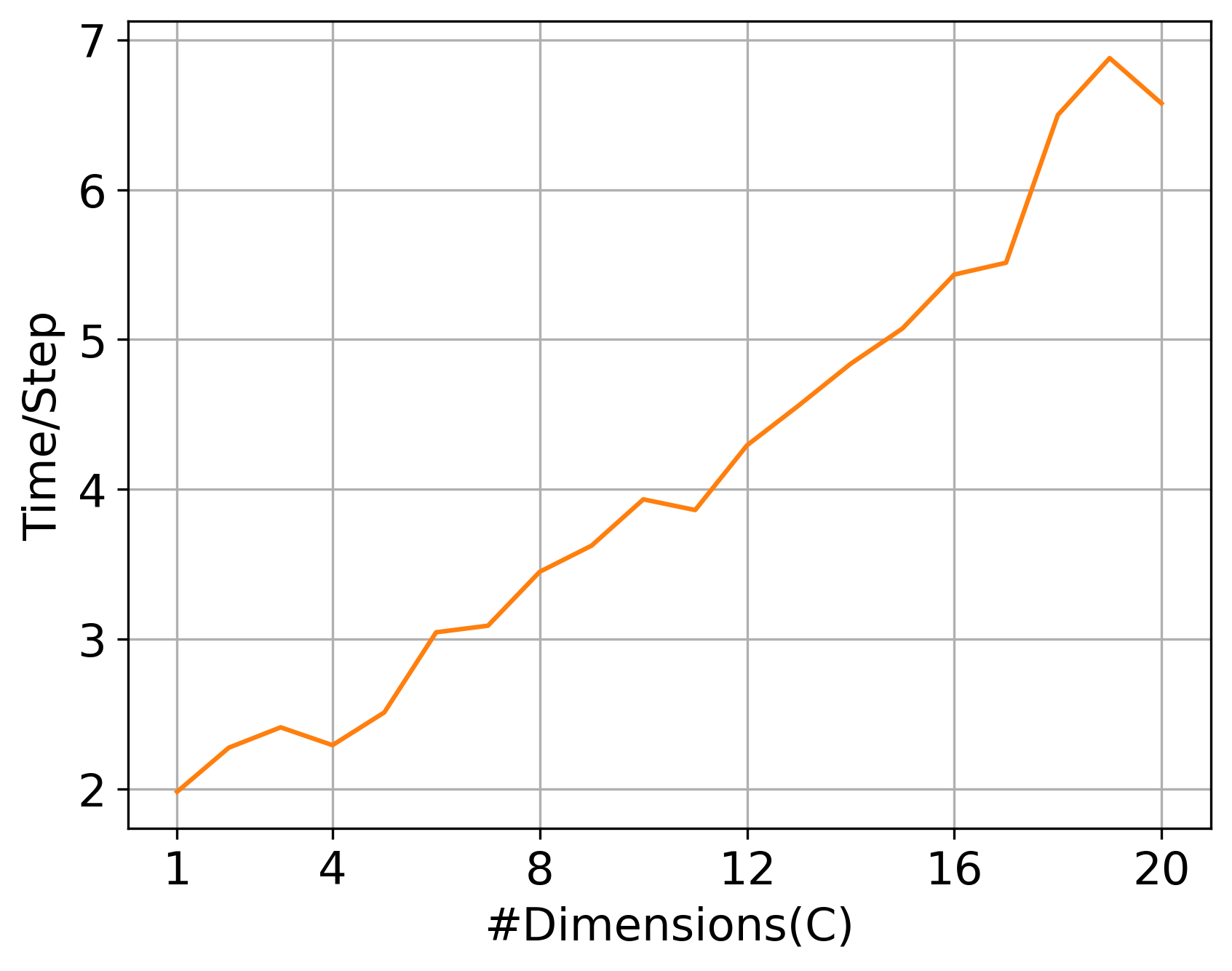}}
\caption{\label{fig:cost} \small{The average computational cost of iLQR optimization performed during each iteration, plotted against the dimension of control subspace $C$.}}
\vspace{-15px}
\end{figure}
\begin{figure}[ht]
\centering
\scalebox{.7}{
\includegraphics[width=\linewidth]{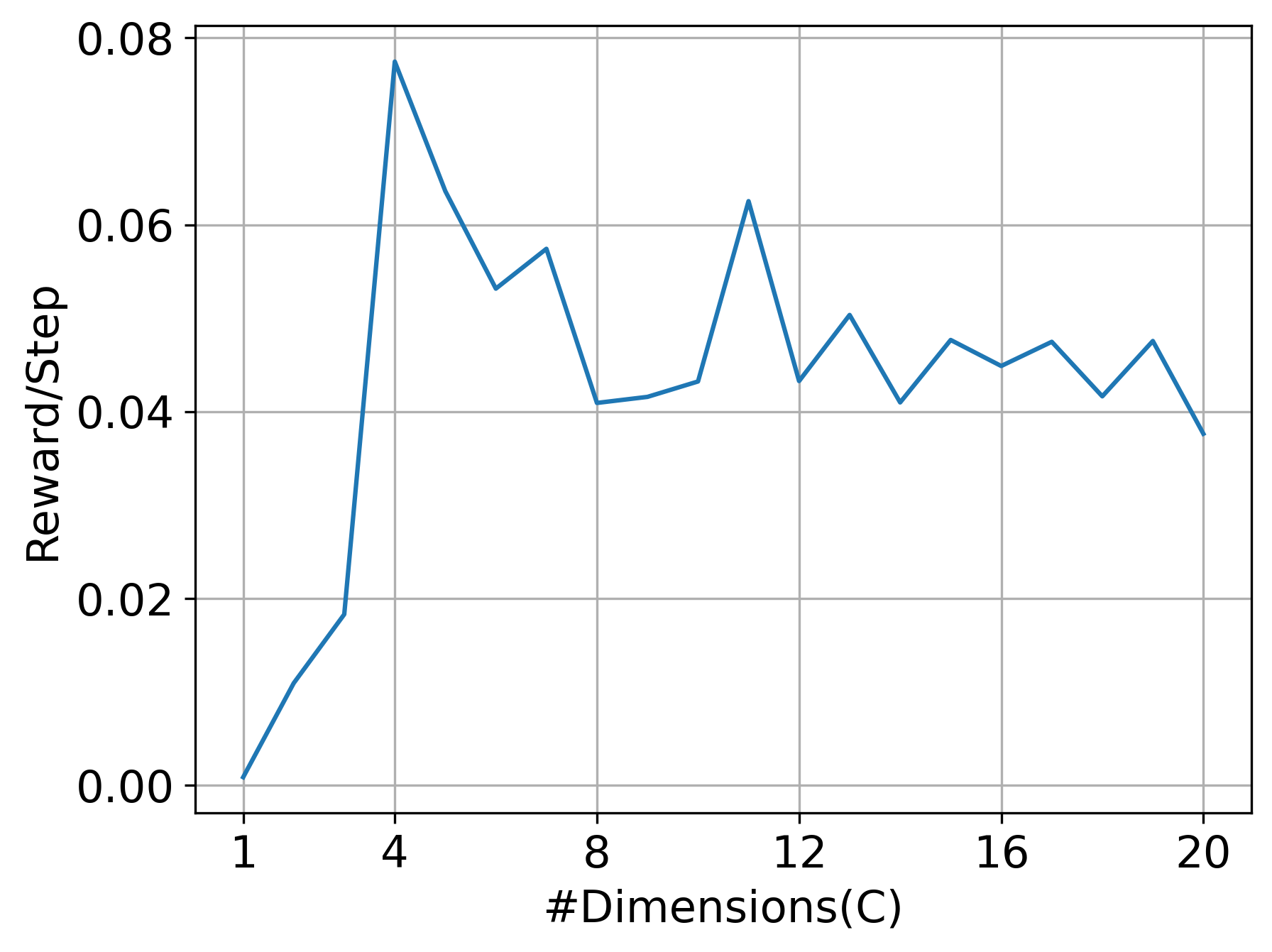}}
\caption{\label{fig:compareC} \small{The performance of controller restricted to the first $C$ bases of $B_r$ for the cross walk task.}}
\vspace{-15px}
\end{figure}
\begin{figure*}[ht]
\centering
\begin{tabular}{ccccc}
\includegraphics[width=.19\linewidth]{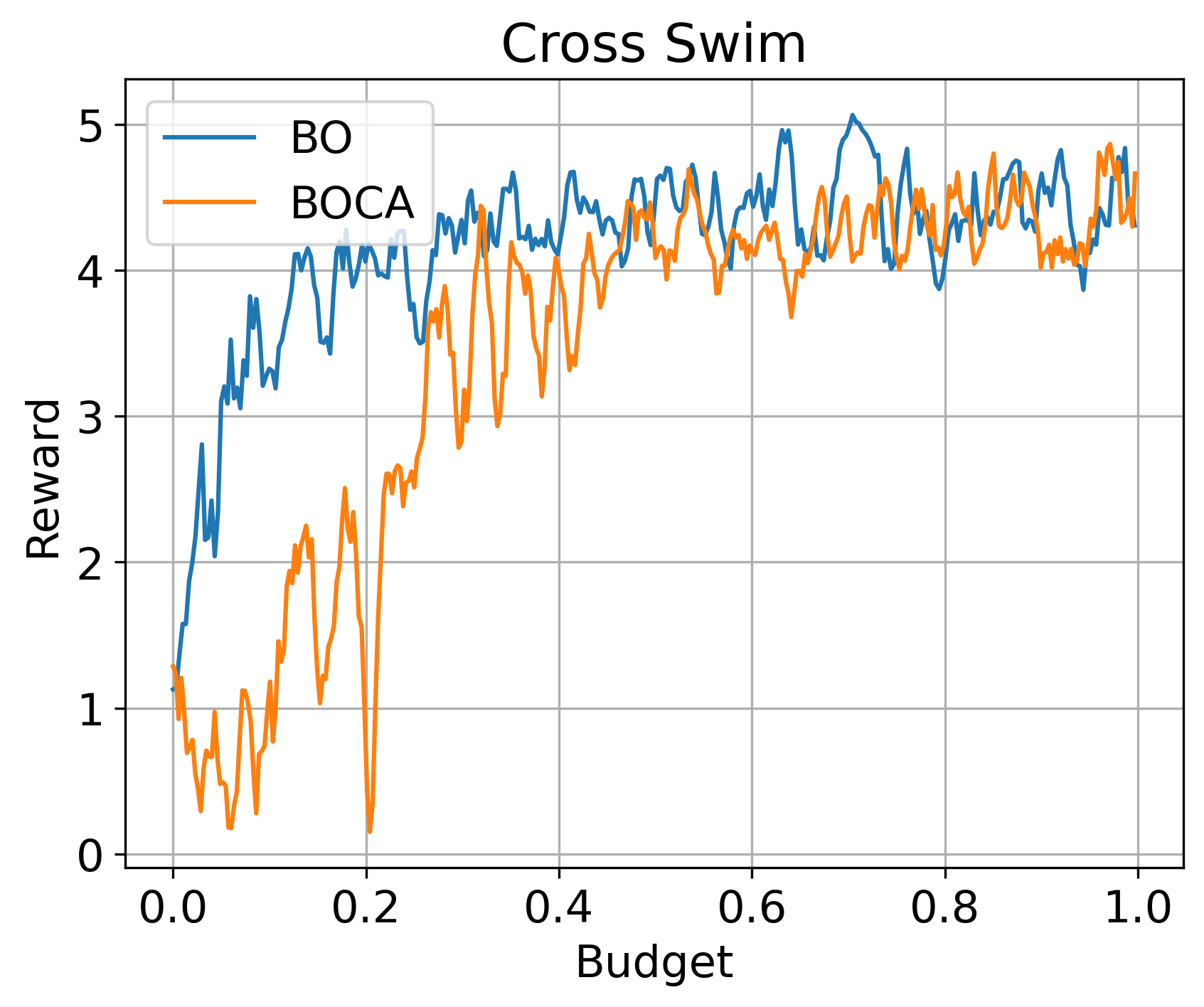}&
\includegraphics[width=.19\linewidth]{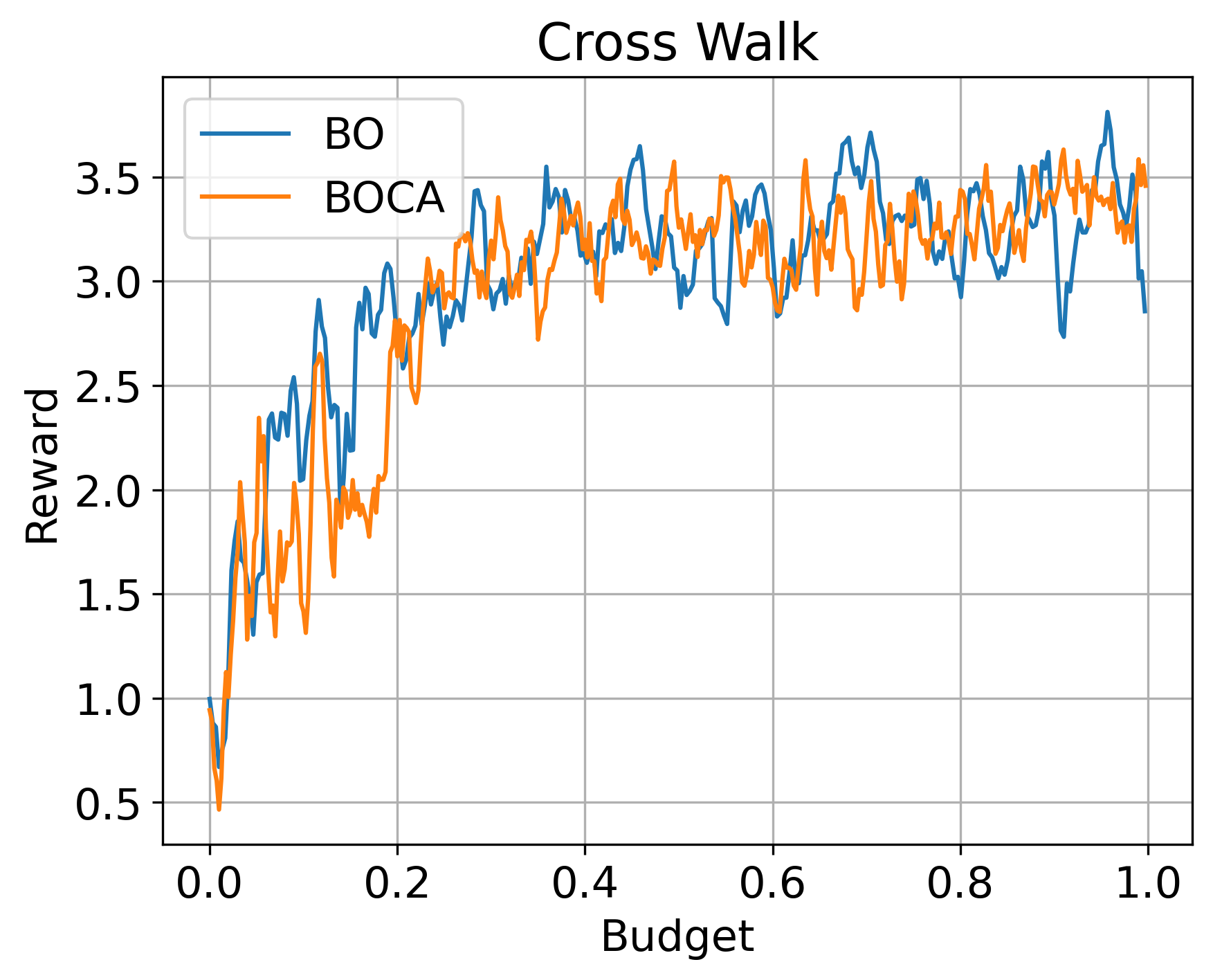}&
\includegraphics[width=.19\linewidth]{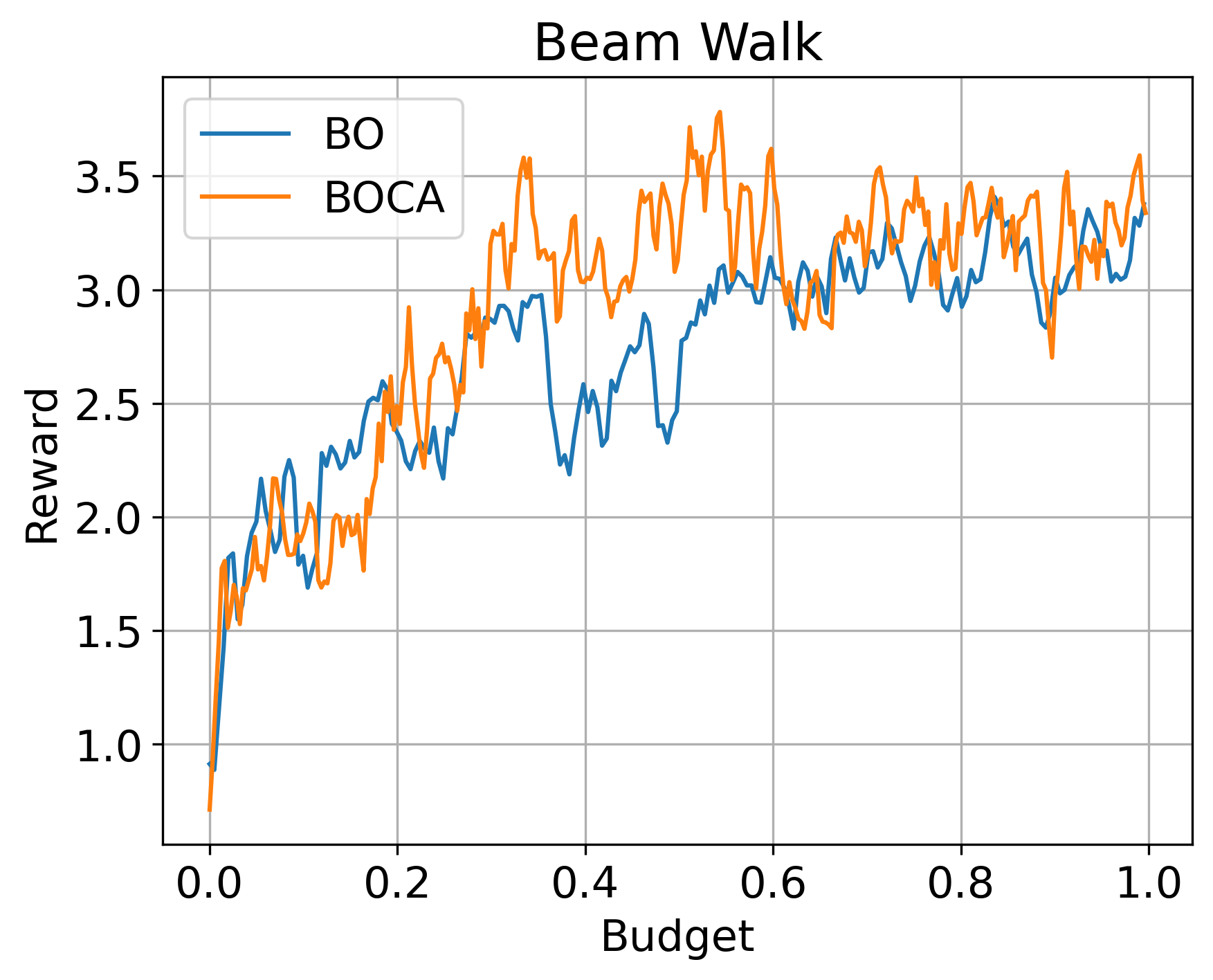}&
\includegraphics[width=.19\linewidth]{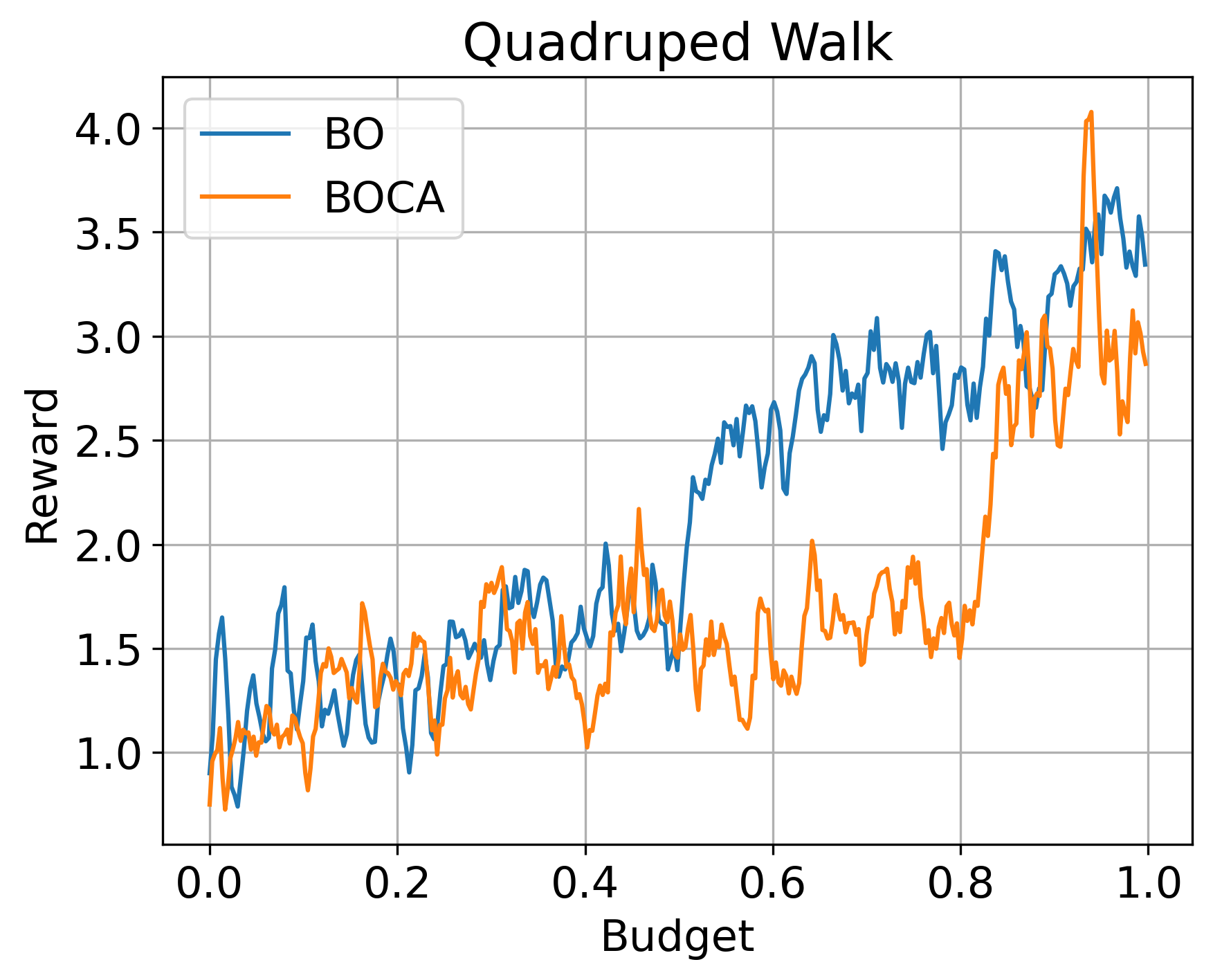}&
\includegraphics[width=.19\linewidth]{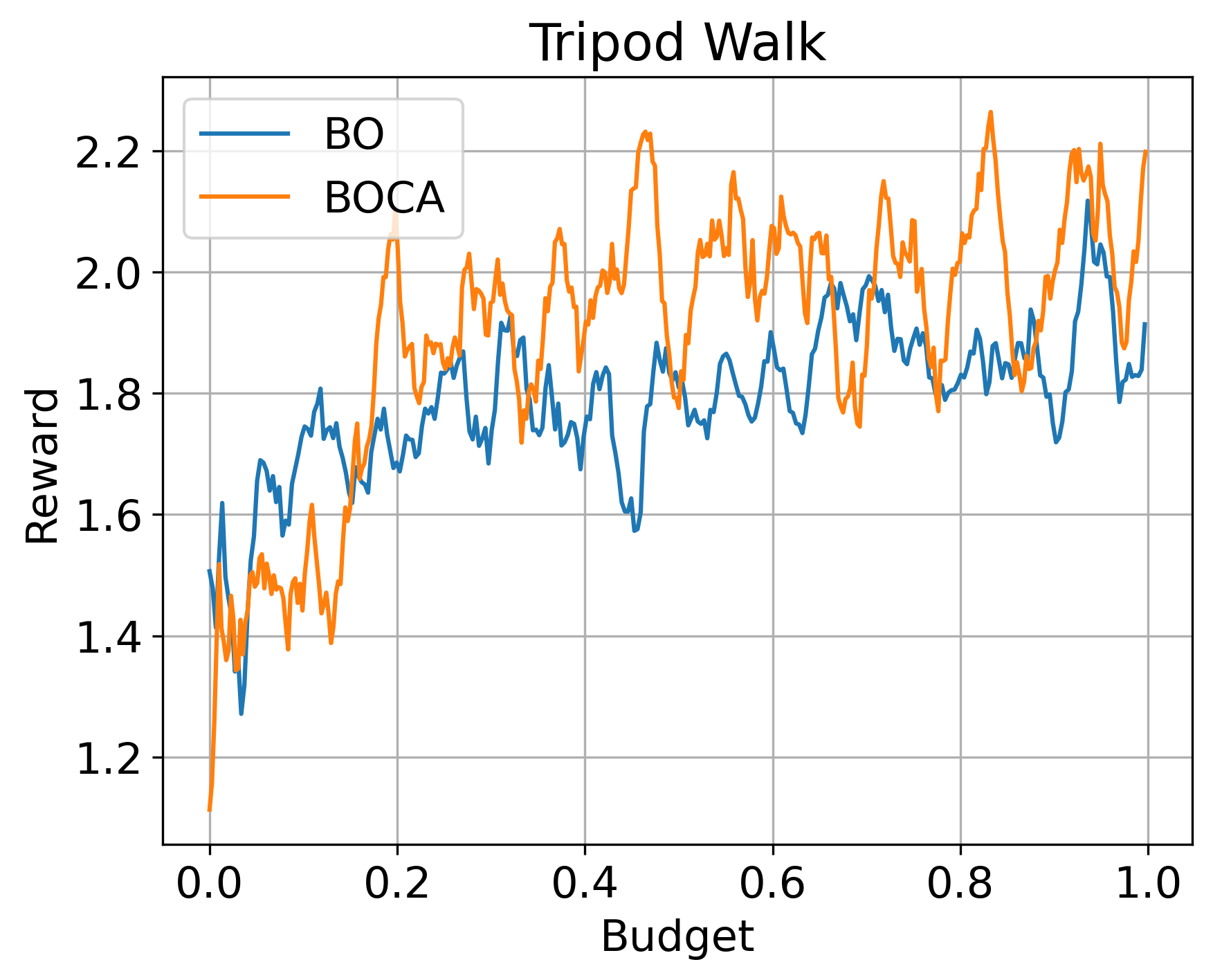}
\end{tabular}
\vspace{-5px}
\caption{\label{fig:BOVsBOCA} \small{The convergence history of BO and BOCA for our five benchmark problems (sampled reward function value at each iteration plotted against computational time).}}
\vspace{-15px}
\end{figure*}
\begin{figure}[ht]
\centering
\scalebox{.65}{
\includegraphics[width=\linewidth]{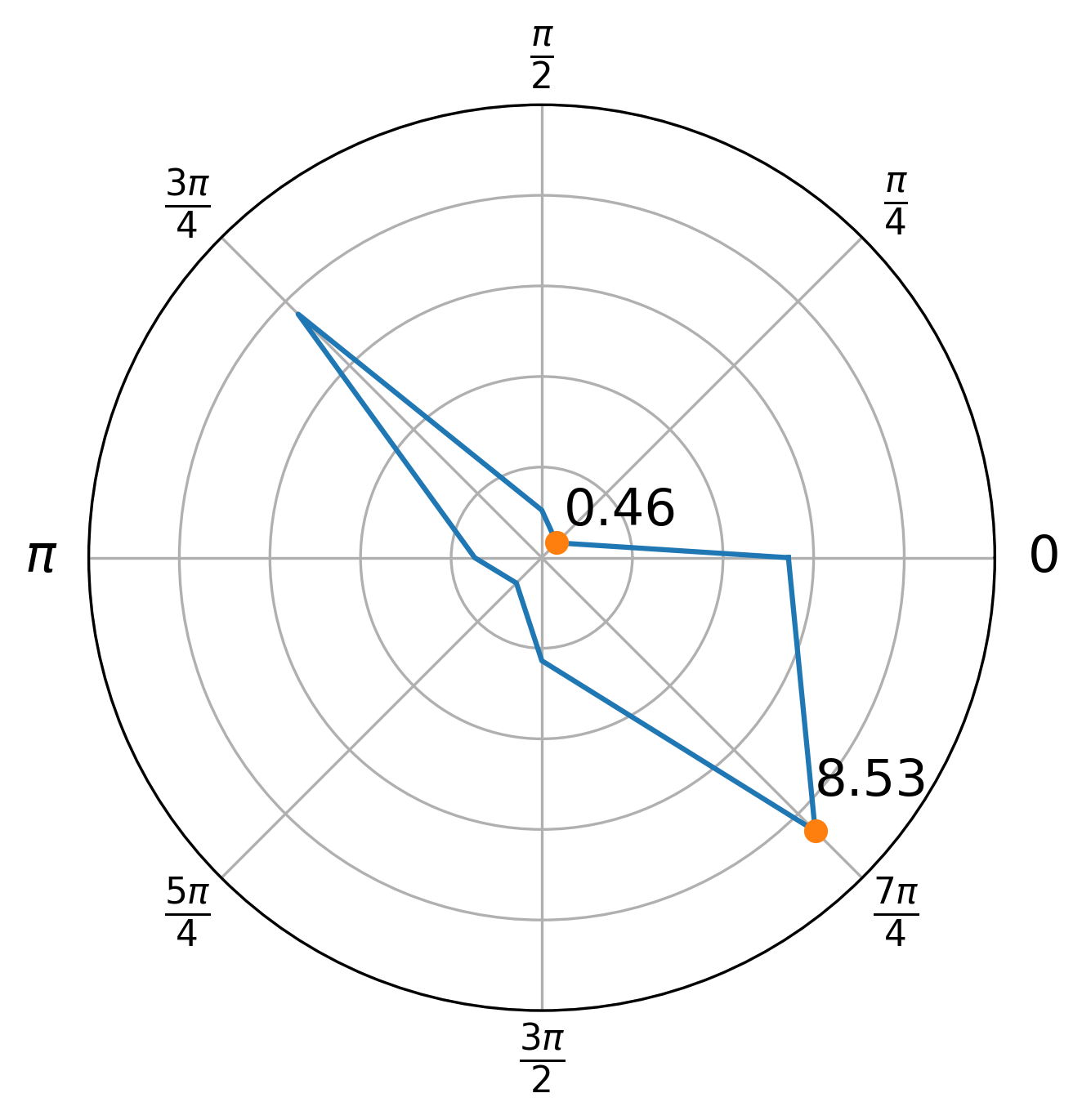}}
\caption{\label{fig:8Dir} \small{For the soft-cross, we run RBBO for the robot to walk along 8 directions. The performance improvement is plotted as a polar graph. Our method brings at least 46.45\% percent and at most 853.50\% of performance improvement.}}
\vspace{-15px}
\end{figure}
\begin{figure}[ht]
\centering
\scalebox{.7}{
\includegraphics[width=\linewidth]{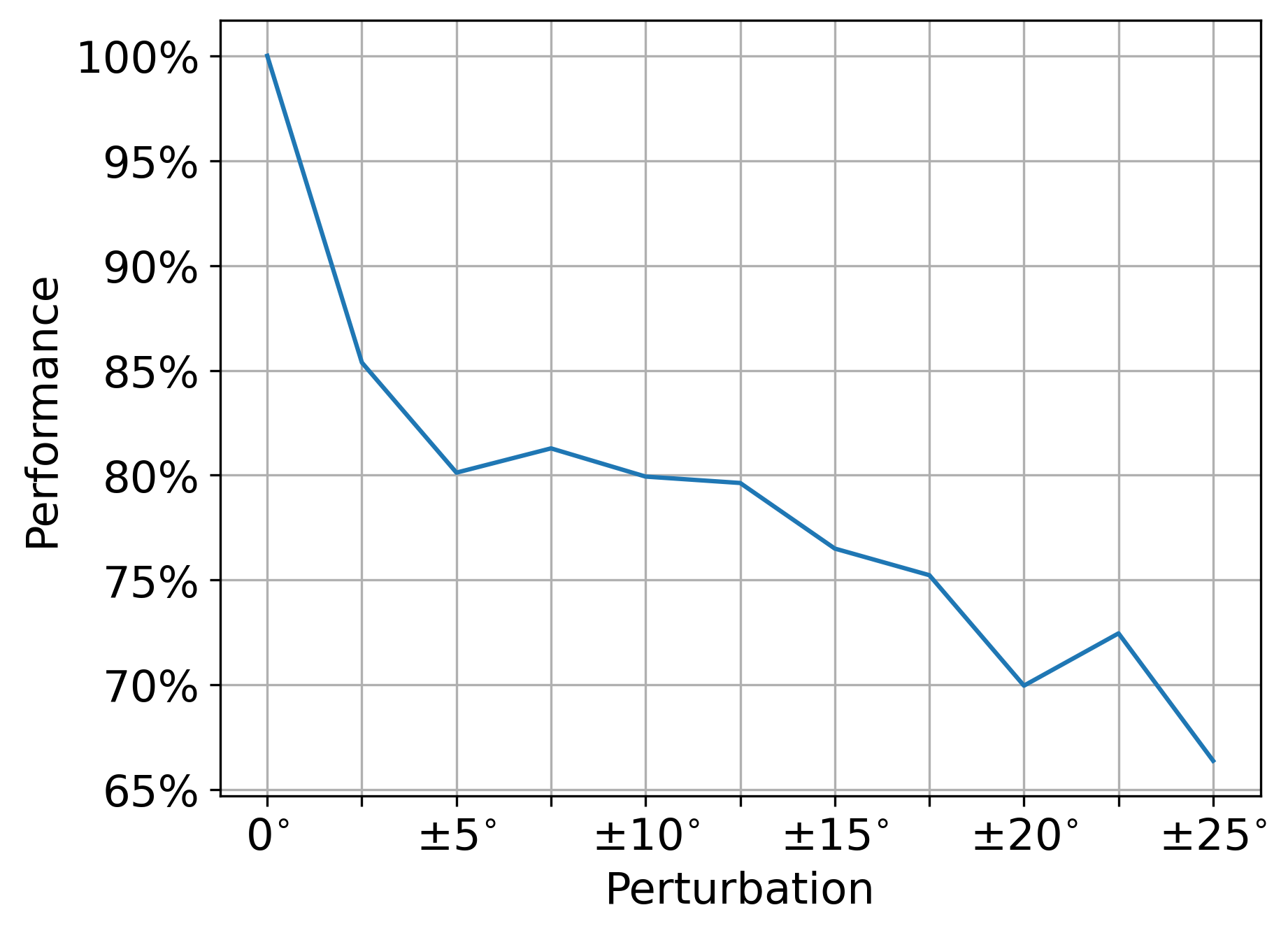}}
\caption{\label{fig:DirPerturb} 
\small{The performance of controller plotted against a perturbation of the walking direction. Our method achieves 80\% performance when the walking direction perturbation is less than $\pm10^\circ$.}}
\vspace{-15px}
\end{figure}
\subsection{Computational Cost Versus Performance}
We first demonstrate the significantly reduced computational cost due to the use of low-dimensional control subspace. In~\prettyref{fig:cost}, we plot the average computational cost of one round of iLQR optimization against the dimension of control subspace $C$. It is not surprising that the cost grows superlinearly with respect to $C$. Indeed, MPC only exhibits nearly interactive performance when $C\leq20$. Although our simulator is not highly optimization, the speedup due to further low-level optimization is limited. Moreover, we show that controlling all the reduced bases is unnecessary. In \prettyref{fig:compareC}, we plot the performance of controller (measured by $\mathcal{R}$) when the control signal is restricted to the first $C$ bases of $B_r$, i.e., $B_c$ is an identity matrix. We see that the performance improvement levels off after the first $4$ dimensions. This observation strongly suggests the use of a low-dimensional control subspace, so we choose to use $C=3$ for all our examples. Further, since the performance of BO can degrade significantly in high-dimensional search spaces, we always limit $M=20$.

\subsection{Performance of BO Versus BOCA}
We run both BO and BOCA for 10 hours for each benchmark and compare their results with the identity baseline, i.e., using the first $C$ bases of $B_r$ by setting $B_c$ to an identity matrix. Note that the identity baseline is already a reasonable initial guess, since it controls the $C$ most salient deformable modes. In~\prettyref{fig:BOVsBOCA}, we profile the performance of RBBO over our five benchmarks problems, by plotting the reward function value sampled at each iteration against computational time for fairness of comparison. In \prettyref{tab:results}, we show that RBBO can improve the controller performance by 544.32\% at most and 229.49\% on average, as compared with the identity baseline. These results essentially imply that the most effective control modes is different from the most salient deformation modes. On the downside, BOCA does not bring a distinguishable benefit over BO. We can run more iterations of BOCA within the same amount of computational time, but the overall reward improvement is comparable. 

\subsection{Task-Sensitivity of Controller}
We finally demonstrate the robustness of our method by analyzing the sensitivity of our method to the change of tasks. To this end, we run RBBO for the soft-cross to walk along 8 different directions. As illustrated in~\prettyref{fig:8Dir}, our method consistently improve the controller performance by at least 46.45\% and at most 853.50\%, as compared with the identity baseline. Further, we evaluate the generalization ability for an optimized $B_c$ to other tasks. To this end, we optimize $B_c$ for the robot to walk along a fixed direction, and we then apply the same $B_c$ to slightly different walking directions by perturbing the desired angle of direction. In this case, the performance is plotted against the angle of perturbation in~\prettyref{fig:DirPerturb}. Our method achieves 80\% performance when the direction perturbation is less than $\pm10^\circ$, as compared with the unperturbed version of our approach, while its performance quickly decreases as the perturbation further increases. We further observe that the controller performance differs drastically with the task, i.e., walking direction. This is presumably because the simulation subspace is task-biased and not involved in our optimization pipeline. The joint optimization of simulation- and control-spaces is beyond the scope of this work.
\section{Conclusion}
We propose a novel controller design and optimization scheme for soft robots. To circumvent the curse-of-dimensionality, we propose a two-stage dimension reduction method. We first use conventional reduced-order modeling tools to find a simulation subspace, we then introduce an even lower-dimensional control subspace. We propose to optimize the control subspace via Bayesian optimization for maximizing the controller performance. Our results show that our proposed scheme achieves a consistent performance improvement for various soft robots with drastically different modalities. As the major limitation of current work, our method cannot learn a universal control subspace that can generalize to all tasks. In the future, we are considering deep learning models that can predict near-optimal control subspaces, given a certain task. Our method further assumes the ROM simulator accurately models the behavior of the soft robot, while the error introduced by a ROM simulator is beyond the scope of this study. In real world robotic applications, it is an essential future work to analyze the ROM simulator error and its impact on the simulation and prediction of soft robot hardware.
\AtNextBibliography{\footnotesize}
\printbibliography
\end{document}